\documentclass[]{spie} 

    \usepackage{standalone}
    
    \usepackage{amsmath,amsfonts,amssymb}
    \usepackage{graphicx}
    \usepackage[colorlinks=true, allcolors=blue]{hyperref}

    \title{Vision Transformers: the threat of realistic adversarial patches}
    
    \author[*a,b]{Kasper Cools}
    \author[c]{Clara Maathuis}
    \author[d]{Alexander M. van Oers}
    \author[e]{Claudia S. Hübner}
    \author[b, f]{Nikos Deligiannis}
    \author[a]{Marijke Vandewal}
    \author[a]{Geert De Cubber}
    
    \affil[a]{Belgian Royal Military Academy, Brussels, Belgium}
    \affil[b]{Vrije Universiteit Brussel, Brussels, Belgium}
    \affil[c]{Open University of the Netherlands, Heerlen, the Netherlands}
    \affil[d]{Netherlands Defence Academy, Den Helder, the Netherlands}
    \affil[e]{Fraunhofer Institute of Optronics, Ettlingen, Germany}
    \affil[f]{imec, Leuven, Belgium}
    
    \authorinfo{Corresponding author: Cools K., e-mail: kasper.cools@mil.be}
    
\begin{document} 
    \maketitle
    
    \begin{abstract}
    The increasing reliance on machine learning systems has made their security a critical concern. Evasion attacks enable adversaries to manipulate the decision-making processes of AI systems, potentially causing security breaches or misclassification of targets. Vision Transformers (ViTs) have gained significant traction in modern machine learning due to increased 1) performance compared to Convolutional Neural Networks (CNNs) and 2) robustness against adversarial perturbations. However, ViTs remain vulnerable to evasion attacks, particularly to adversarial patches, unique patterns designed to manipulate AI classification systems. These vulnerabilities are investigated by designing realistic adversarial patches to cause misclassification in person vs. non-person classification tasks using the Creases Transformation (CT) technique, which adds subtle geometric distortions similar to those occurring naturally when wearing clothing. This study investigates the transferability of adversarial attack techniques used in CNNs when applied to ViT classification models. Experimental evaluation across four fine-tuned ViT models on a binary person classification task reveals significant vulnerability variations: attack success rates ranged from 40.04\% (google/vit-base-patch16-224-in21k) to 99.97\% (facebook/dino-vitb16), with google/vit-base-patch16-224 achieving 66.40\% and facebook/dinov3-vitb16 reaching 65.17\%. These results confirm the cross-architectural transferability of adversarial patches from CNNs to ViTs, with pre-training dataset scale and methodology strongly influencing model resilience to adversarial attacks.
    
    \end{abstract}
    
    
    \section{Introduction}
    \label{sec:intro} 
    The rising adoption of vision-based AI (Artificial Intelligence) models in border security and law enforcement applications such as border-control kiosks, body-worn cameras, and airborne surveillance platforms \cite{zhang2024MorphingAttack} has increased the attack space on these models which directly implies implications and consequences in relation to their robustness and reliability \cite{Zhou2025VitAttacks10, nguyen2023physical}. In these operational settings an evasion attack does not just degrade accuracy metrics, but it could further imply that illicit goods or hostile actors pass undetected, undermining border security and law-enforcement operations and, by extension, have an impact on national security. This threat is amplified by the physical realisation of adversarial patches, i.e., printable patterns that compel a detector to ignore or mislabel a target even under variable illumination and viewing angles \cite{brown2017aurko,5adbb35b1d3a411b8c2ecf9075e9e38b,pmlrv80athalye18b,Thys_2019_CVPR_Workshops,10.1007/978-3-030-58548-8_1, 10.1117/12.3031144}. Recent models are relying on state-of-the art transformer-based vision models such as Vision Transformers (ViTs) and multimodal hybrids such as CLIP (Contrastive Language-Image Pre-Training) \cite{ali2023clip, radford2021learning}, and now dominate person-detection, entity detection, and re-identification benchmarks in this domain \cite{Ye2024TransformerReIDSurvey, Bayraktar2025ReTrackVLM} as they replace previous convolutional backbones used in different field systems.

    Specifically, a transformer-based vision model splits an input image into a grid of fixed-size patches, then processes the resulting token sequence with the same self-attention layers that power language transformers. By learning long-range relationships among patches rather than local convolutional filters, ViTs build a global representation of the scene that can feed downstream tasks such as classification, detection, or segmentation \cite{Khan2022}. At the same time, CLIP extends this idea to two modalities: it trains a vision encoder and a text encoder jointly so that images and their paired textual captions map to nearby points in a shared embedding space. After pre-training on a large set of image and text pairs, CLIP can rank how well it aligns with candidate text prompts making it a versatile backbone for multimodal applications\cite{Pan2022} and an appealing target for both creative use and adversarial exploitation. Their global self-attention and token mixing provide both a superior recognition accuracy and a degree of resilience to pixel-level noise \cite{Ebert2023PLGViT}. Nevertheless, recent research reveals that ViTs remain susceptible to structured, patch-based perturbations \cite{Tian2024ViTRobustness}.
    
    The aim of this study is to analyse and explain the susceptibility of state-of-the-art ViTs and CLIP models to realistic, physical-world evasion attacks that suppress person-detection outputs. Focusing on security deployments present in border security and law enforcement domains, the scope is delimited to patch-based attacks that can potentially be printed on garments or equipment, generalised across viewpoints, and remain functional under natural deformations. On this behalf, a Creases Transformation (CT) layer\cite{Guesmi2024} is introduced to the adversarial patch pipeline developed for CNN detectors by applying physically plausible geometric distortions (folds and local stretching) to each candidate patch during optimisation. To this end, craft patches are applied on a YOLOv5-CNN baseline, further transferred to ViT-based person detectors  in order to measure the success rates. The results obtained show the effectiveness of the approach taken and provide a reproducible testbed for future defensive research. This further motivates security oriented data-augmentation pipelines that inject wrinkle-induced shape noise during training, thereby immunising detectors to the very distortions exploited by attackers. At the same time, this research provides an evaluation protocol based on a reproducible benchmark for building new defence mechanisms under conditions that mirror real operational environments and conditions, accelerating the iterative hardening of vision systems deployed in critical security domains.
    
    The outline of this article is structured as follows. Section \ref{sec:related} discusses relevant research studies conducted in this domain. Section \ref{sec:method} presents the methodological approach taken and the dataset used together with design and implementation choices taken in this process. Section \ref{sec:results} reflects on the results obtained for the experiments conducted and provides a series of recommendations from a defensive perspective. At the end, Section \ref{sec:conclusion} discusses concluding remarks and future research perspectives. 
    
    \section{Related Research}
    \label{sec:related}
    ViTs and hybrid multimodal models such as CLIP are increasingly being used in mission-critical systems and applications such as autonomous surveillance, precision-guided munitions, and human-machine teaming. This happens due to various advantages that they provide, e.g., superior accuracy and increased robustness when compared to their convolutional counterparts \cite{naseer2021intriguing, tu2025toward}. Nevertheless, recent studies reflect the fact that small, printable adversarial patches like G-Patch can universally mislead ViTs across viewing angles \cite{shao2023}, while AdvCLIP shows that a single cross-modal patch can corrupt every downstream task that inherits a compromised CLIP encoder \cite{zhou2023advclip}. 
    
    At the same time, dynamic variants that embed creases and fabric distortion further demonstrate physical plausibility reflecting the operational relevance of these attacks \cite{Guesmi2024, gu2022vision}. As failures in these contexts translate directly into missed targets or collateral damage by the systems used, characterising the attack surface of ViT-based systems and understanding why CNN-derived defenses such as patch excision or feature regularisation only partially transfer represents a prerequisite for designing resilient and robust architecture-aware countermeasures suitable for defence and security deployments. Further, in token-based attacks, the unique patch-token processing of ViTs is exploited showing that even a single adversarially manipulated token can significantly disrupt model predictions, a vulnerability less pronounced in CNNs \cite{Joshi2025}. Through a transfer mechanism, in transfer-based attacks, adversarial examples on a surrogate model are generated and transferred to a target model. This mechanism proves to be highly effective against ViTs especially when the attack methods are adapted to the self-attention mechanisms inherent in these architectures \cite{zhang2023, wei2022towards}. 
    
    Besides adversarial attacks, ViT are vulnerable to backdoor attacks, where an adversary poisons a portion of the training data with triggers that, when present at inference, can cause the model to misclassify inputs. On this behalf, Subramanya et al. (2022)\cite{subramanya2022} found that ViTs are at least as vulnerable to backdoor attacks as CNNs, as the attention mechanisms are playing a significant role in the propagation and detection of backdoor trigger. Moreover, representation-level attacks, which target the internal embeddings of ViTs, revealed that adversarial perturbations can propagate through the self-attention layers, leading to significant changes in the model's internal representations and final outputs \cite{islam2025, mahmood2021robustness}.
    
    In the security domains, adversarial patches, either physical objects or printed pattern, can be strategically placed within a scene to evade detection or trigger false alarms in automated border control or law enforcement systems, undermining the reliability of these critical infrastructures \cite{Fawole2025}. To these are added the transferability and practicability of these attacks in various operational settings since adversarial patches crafted for one ViT model can often transfer to other architectures, including those used in robust or defended settings \cite{alam2025adversarial, ming2024boosting}. These studies show the need for robust and adaptive ViT-specific defense mechanisms in security-critical applications to prevent adversarial attacks from compromising border security and law enforcement operations.

    \section{Method}
    \label{sec:method}
    This study investigates the vulnerability of transformer-based image classifiers to adversarial patch attacks. Four pre-trained Vision Transformer (ViT-B/16) models were fine-tuned on a custom binary classification task (person vs. non-person). The models, along with their pre-training strategies and fine-tuning results, are summarised in Table~\ref{tab:models}. Notice the slight underperformance of dinov3 compared to the other finetuned models. Because this research finetunes all models using the same sample dataset that could attribute to a slightly lower F1 score at this stage. Given the number of parameters for dinov3, using a larger sample dataset could be considered to give the model room to learn more intricate features. However, for the purpose of this research the same dataset was used for consistency.

     \begin{figure}[htbp]
        \centering
        \includegraphics[width=\textwidth]{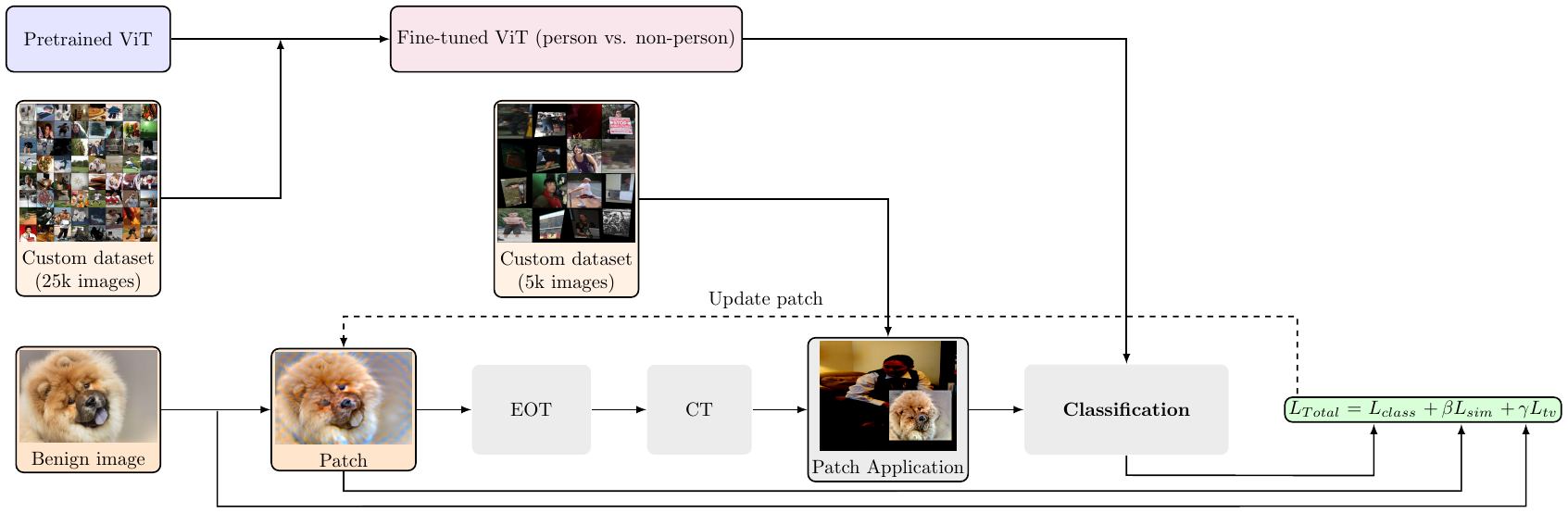}
        \caption{Overview of the fine-tuning and patch generation.}
        \label{fig:overview}
    \end{figure}
    
    \subsection{Dataset Composition}
    To evaluate the transferability of adversarial patch attacks on a classification task, a suitable person classification dataset is required. However, existing person-related datasets are predominantly designed for detection tasks, providing bounding box annotations rather than image-level classification labels. Furthermore, given that ViT architectures typically require substantial amounts of training data for effective fine-tuning, the limited size of available person classification datasets posed an additional limitation. To address these, a custom binary classification dataset was composed specifically for person and non-person classification.
    
    The dataset for this research was derived from the Microsoft COCO dataset, which provides annotations for 80 different object categories spanning people, animals, vehicles, furniture, and various other objects \cite{COCO2014}. Utilising the train2017 subset, a custom filtering script was developed based on COCO's existing annotations to create two distinct object categories: person vs non-person classes. For person images, images containing between 1 and 3 person annotations per image were retained.
    
    This constraint ensured manageable complexity while maintaining sufficient contextual diversity for training robust classification models. Non-person images were selected from those containing no person class annotations, encompassing the full range of COCO's remaining object categories including vehicles, animals, household objects, and outdoor scenes.
    
    Each qualifying image was processed to extract object instances using bounding boxes with an additional 15\% padding to preserve contextual information for training. The final dataset comprised of 25000 images: 20000 for training, 2500 for validation, and 2500 for testing, ensuring class balance across all subsets.
    
    In addition to ViT fine-tuning, this dataset was employed for adversarial patch generation following an adapted methodology established in the DAP framework \cite{Guesmi2024}. For this purpose, a reduced subset of 5000 images was randomly selected from the training set, providing sufficient diversity for effective adversarial training.
    
    \subsection{Fine tuning ViT's}
    \subsubsection{Baseline Model Training}
    To establish robust target models for adversarial evaluation, four pre-trained ViT models were finetuned on our custom binary person vs. non-person classification dataset. The selection of these specific models provides comprehensive coverage of different pre-training paradigms: self-supervised learning on ImageNet-1k (\texttt{facebook/dino-vitb16}), and supervised learning on both the larger ImageNet-21k dataset\\
     (\texttt{google/vit-base-patch16-224-in21k}), and on the smaller ImageNet-1k (\texttt{google/vit-base-patch16-224}).

    \subsubsection{Layer Freezing Strategy}
    A selective layer freezing strategy was employed to preserve pre-learned representations while enabling task-specific adaptation. Specifically, the patch embedding and positional embedding layers, as well as encoder layers 0--7 (including their associated LayerNorm parameters) were frozen. This configuration preserves the low-level feature representations learned during pre-training while allowing the upper encoder layers (8--11) and the classification head to adapt to the binary classification task. This approach balances computational efficiency with effective transfer learning.
    
    \begin{table}[h]
    \centering
    \caption{Performance of fine-tuned ViT models on the person vs. non-person classification task.}
    \label{tab:models}
    \begin{tabular}{lcccc}
    \textbf{Model} & \textbf{Accuracy} & \textbf{F1 Score} & \textbf{Precision/Recall} \\ 
    \hline
    google/vit-base-patch16-224 &  99.60\% & 99.62\% & 99.54\% / 99.70\% \\ 
    google/vit-base-patch16-224-in21k & 99.60\% & 99.61\% & 99.54\% / 99.70\% \\ 
    facebook/dino-vitb16 & 96.72\% & 97.40\% & 95.13\% / 99.17\% \\ 
    facebook/dinov3-vitb16 & 81.00\% & 78.20\% & 91.21\% / 68.58\%
    \end{tabular}
    \end{table}
    
    \subsubsection{Training Configuration}
    The fine-tuning process employed hyperparameters selected based on preliminary experiments and established best practices for ViT fine-tuning. All models were trained with a batch size of 128 and a fixed learning rate of $0.0001$ using the Adam optimiser. All experiments were conducted on dual NVIDIA RTX 6000 Ada Generation GPUs (48 GB VRAM each) for  training.
    
    Standard data augmentation techniques were applied during training, including random horizontal flipping (probability = 0.5) and colour jittering (brightness = 0.1, contrast = 0.1) to enhance model generalisation. Model performance was evaluated using multiple metrics including accuracy, precision, recall, and F1 score to ensure comprehensive assessment of classification performance.
    
    \subsection{Adversarial Patch Generation Framework}
    This research adapts the DAP (Dynamic Adversarial Patches) framework from object detection on CNN-based architectures to classification tasks using ViTs, maintaining core principles while implementing modifications for transformer architectures. The adapted pipeline is depicted in Figure \ref{fig:overview}.

    \subsubsection{Training Data Selection}
    Adversarial patch optimisation utilised a subset of 5,000 images that were not present in the training set. This subset size represents a balance between computational constraints and sufficient diversity for effective adversarial training, following established practices in adversarial patch research.
    
    \subsubsection{Expectation over Transformation}
    To enhance patch robustness and physical realisability, the following Expectation over Transformation methods \cite{pmlr-v80-athalye18b} were implemented and applied during adversarial optimisation:
    \begin{itemize}
        \setlength\itemsep{0.2em}
        \item \textbf{Random Rotation:} Patches undergo random rotation within $\pm20^\circ$ to simulate varying orientations in real-world scenarios.
        \item \textbf{Dynamic Resizing and Scaling:} Patch dimensions are dynamically adjusted between 25\% and 125\% of the original size to account for varying distances and object scales.
        \item \textbf{Affine Transformations:} Random affine distortions with shear parameters of $\pm0.7$ simulate perspective changes and viewing angle variations.
        \item \textbf{Brightness and Contrast Adjustment:} Brightness jittering of $\pm0.1$ and contrast scaling within [0.8, 1.2] replicate diverse lighting conditions.
        \item \textbf{Noise Injection:} Gaussian noise with standard deviation 0.1 is added to enhance robustness against environmental variations.
    \end{itemize}
    
    \subsubsection{Crease Transformation Block Adaptations}
    \begin{itemize}
        \item \textbf{Creases Transformation (CT):} This transformation applies directional quadratic displacement fields to generate smooth spatial distortions, simulating realistic fabric deformations when patches are applied to clothing or flexible surfaces.
    \end{itemize}

    \subsubsection{Patch Training Configuration}
    Patches were initialised as 128×128 pixel squares, representing approximately 30\% of the input image area, and clipped to a maximum size of 60\% of the input image during optimisation. The placement strategy involved random positioning of patches on images during both training and testing phases to ensure robustness across different spatial locations. Training proceeded for model-specific optimisation steps (detailed in Table~\ref{tab:results}) using the Adam optimiser with a learning rate of 0.001. Each step a single crease is added to the patch generated after being processed through the EOT block. The vantage point is chosen randomly within the image pane and the direction is determined based on a randomly selected angle within a range of 5 degrees. During evaluation, optimised patches were randomly positioned and resized to 30-60\% of the original image size. All experiments used random seed 2 for initialisation and transformation sampling to ensure reproducibility.
        
    \subsection{Loss functions}
    
    Our adversarial patch optimisation follows the same methodological framework established in the DAP methodology proposed by Guesmi et al. (2024) \cite{Guesmi2024}. In order to make this method work with the approach in this research, the formula has been slightly adapted for ViT classification tasks. The total loss function maintains the same structure:
    
    \begin{align}    
        L_{Total} & = L_{class} + \beta L_{sim} + \gamma   L_{tv}
    \end{align}    
    where $\beta$ and $\gamma$ are weighting hyperparameters.\\
    
    The key modifications to the original DAP implementation lies in the primary adversarial loss function. While the original paper targeted object detection models, our classification-focused approach requires a different formulation. Additionally, this research found that the weighting parameter $\alpha$ for the detection loss in the original formulation did not affect the overall calculation and was therefore removed.
    
    \textbf{Classification Loss}: Adapted from the original detection loss $L_{det}$, our classification loss $L_{class}$ targets the binary person vs. non-person prediction:
    
    \begin{align}
        L_{class} & = -\frac{1}{n} \sum_{i=1}^{n} C_i
    \end{align}
    where $C_i$ represents the predicted probability for the person class on the i-th sample. This approach encourages misclassification by minimising the correct class probability. \\
    
    \textbf{Preserved functions}: The similarity loss $L_{sim}$ and total variation loss $L_{tv}$ remain unchanged from the original DAP formulation \cite{Guesmi2024}:
    \begin{align}
        L_{sim} & = -\left(\frac{\sum_{i,j}P_{i,j}N_{i,j}}{\sqrt{\sum_{i,j}P_{i,j}^2}\sqrt{\sum_{i,j}N_{i,j}^2}}\right)^2 \\
        L_{tv} & = \sum_{i,j}\sqrt{(P_{i+1,j} - P_{i,j} + (P_{i,j+1} - P_{i,j})^2}
    \end{align}
    where $P_{i,j}$  represents the patch pixel values and $N_{i,j}$ represents the original pixel values.
    
    Following the original DAP framework's approach, this research maintained the hyperparameter values $\beta = 4$ and $\gamma = 0.5$. These values proved to be effective for the specific case of this research as well.
    
    \section{Results}
    \label{sec:results}
    This section presents the experimental results evaluating the transferability of adversarial patch attacks from CNN-based architectures to ViT classification models. The effectiveness of patches generated using the adapted DAP framework is analysed across the four different ViT variants. 
    
    \begin{figure}
        \centering
        \begin{minipage}{0.3\linewidth}
            \centering
            \includegraphics[width=\linewidth]{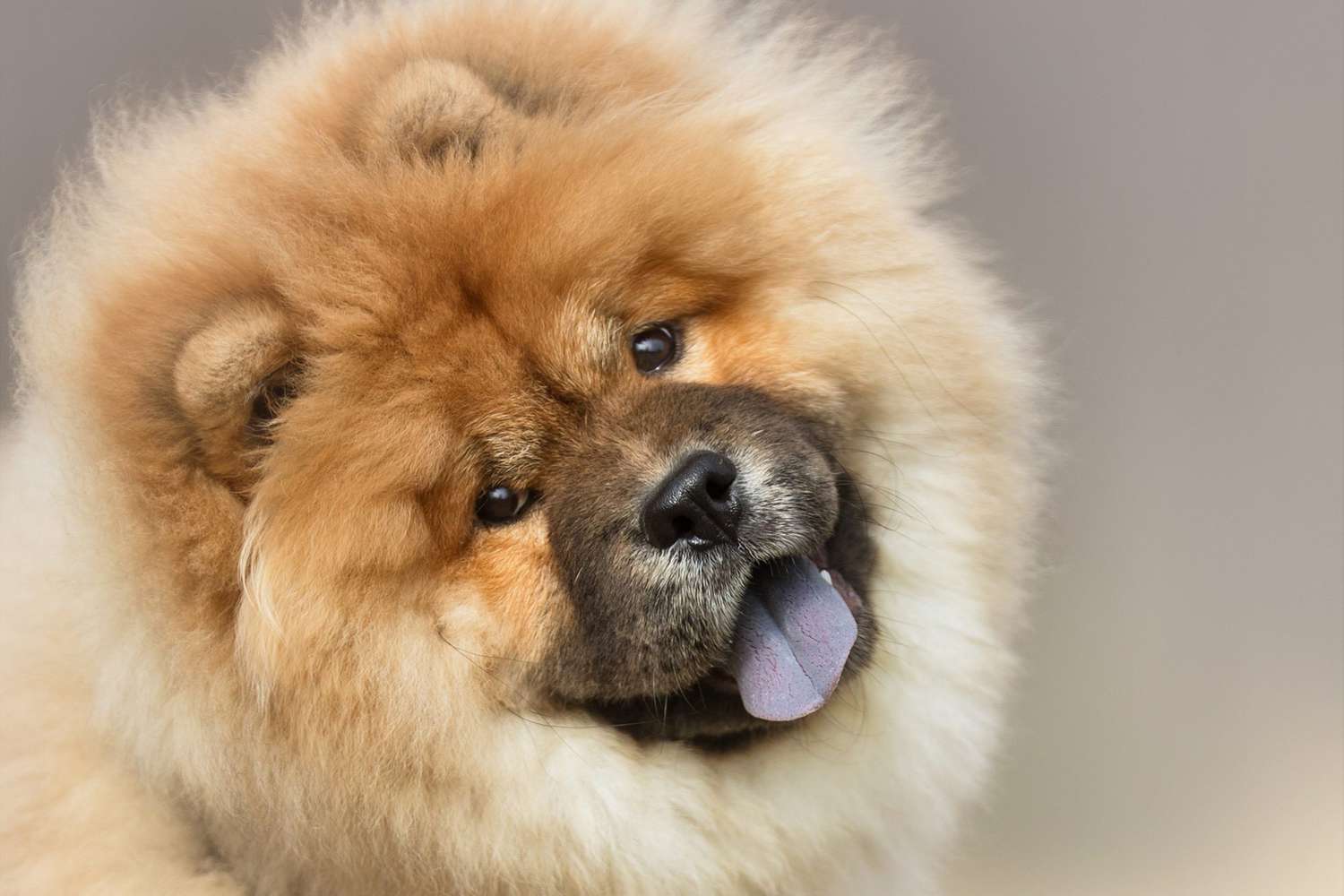}
        \end{minipage}
        \hspace{0.05\linewidth}
        \begin{minipage}{0.3\linewidth}
            \centering
            \includegraphics[width=\linewidth]{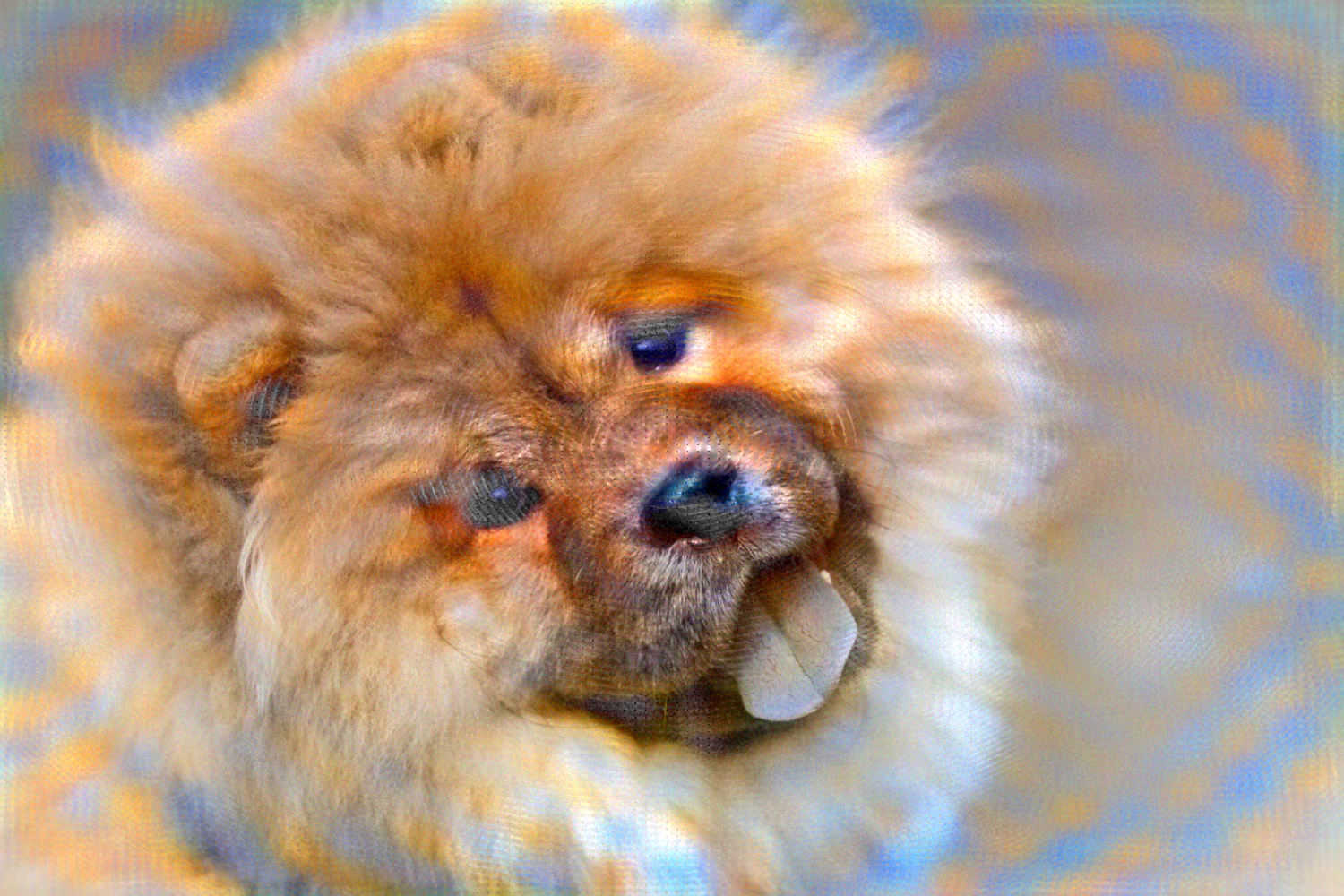}
        \end{minipage}
        \caption{Left: Original image. Right: Resulting adversarial patch}
        \label{fig:sidebyside}
    \end{figure}
    
    The transferability of adversarial patches across different ViT architectures was evaluated by measuring Attack Success Rate (ASR), which is defined as
    \begin{equation}
        ASR = \frac{\mbox{number of false negative for the class `person'}}{\mbox{total number of images with class `person'}}
    \end{equation}
    
    ASR is defined as the proportion of test images with class `person` where patches successfully cause misclassification (i.e., false negatives), which corresponds to the miss rate under attack \cite{Tao2023}.
    
    \begin{table}[htb]
        \caption{Patch effectiveness and training details for different ViT models}
        \label{tab:results}
        \begin{tabular}{lcccc}
        \textbf{Target Model} & \textbf{Accuracy} & \textbf{Accuracy with Patch} & \textbf{ASR} & \textbf{Training Steps} \\ \hline
        google/vit-base-patch16-224 & 99.40\% & 33.40\% & 66.40\% & 4015 \\ 
        google/vit-base-patch16-224-in21k & 99.40\% & 59.60\% & 40.04\% & 3024 \\
        facebook/dino-vitb16 & 95.80\% & 0.02\% & 99.97\% & 3039 \\ 
        facebook/dinov3-vitb16 & 94.20\% & 32.80\% & 65.17\% & 5328
        \end{tabular}
    \end{table}
    
\begin{figure}
        \centering
        \begin{minipage}{0.3\linewidth}
            \centering
            \includegraphics[width=\linewidth]{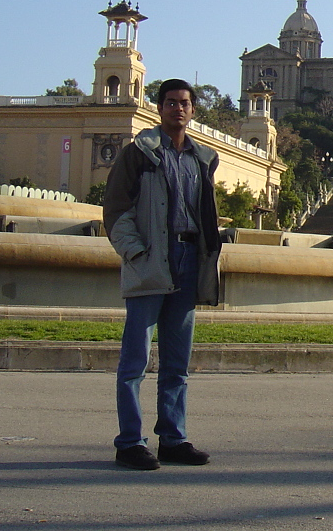}
        \end{minipage}
        \hspace{0.05\linewidth}
        \begin{minipage}{0.3\linewidth}
            \centering
            \includegraphics[width=\linewidth]{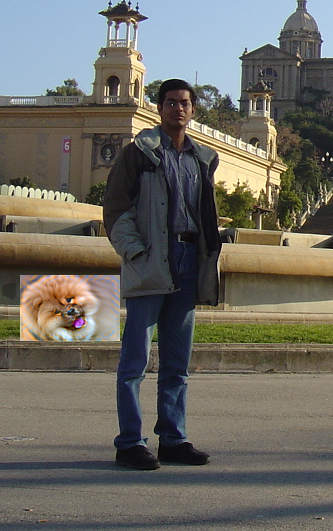}
        \end{minipage}
        \caption{Left: Original image (99.35\% probability). Right: Patched image (30.28\% probability)}
        \label{fig:original_patched_image}
    \end{figure}

    The generated adversarial patches, as depicted in Figure~\ref{fig:sidebyside}, maintained visual similarity while achieving significant attack success rates across the tested ViT architectures. Notably, the facebook/dino-vitb16 model demonstrated the highest vulnerability with a 99.97\% ASR, followed by the google/vit-base-patch16-224 at 66.40\% ASR. This demonstrates that adversarial patches, originally designed for CNNs, can effectively transfer to ViTs, with some architectures being more vulnerable than others. Figure~\ref{fig:original_patched_image} illustrates the practical application of an adversarial patch generated for the vitb-in21k model when applied to the vitb architecture. This example not only demonstrates the visual implementation of our patch methodology but also hints at the potential for cross-model transferability.

    The attack success rates reveal a correlation with pre-training data scale and methodology. The google/vit-base-patch16-224-in21k model, trained on ImageNet-21k, showed substantially improved robustness (40.04\% ASR) compared to its ImageNet-1k counterpart (66.40\% ASR). This lower ASR strongly suggests that exposure to larger and more diverse datasets during pre-training enhances the model's resilience to these types of adversarial patches.
    The facebook/dino-vitb16 model exhibited the highest vulnerability (99.97\% ASR) despite being trained with self-supervised learning objectives. This result suggests that the original DINO's knowledge distillation approach may have inadvertently made the model more susceptible to localised adversarial perturbations. In contrast, the facebook/dinov3-vitb16 model demonstrated improved robustness (65.17\% ASR), indicating that the architectural and training improvements in DINO v3 successfully address many of the vulnerabilities present in the original formulation.
    The CT-augmented patches exhibited natural-looking creases and maintained visual coherence across all tested models, preserving overall appearance while introducing subtle deformations. However, the dramatic variation in attack success rates—from 40.04\% to over 99\%—reveals that different pre-training strategies and data scales potentially create fundamentally different vulnerability profiles in vision transformers. 
    
    \section{Conclusion}
    \label{sec:conclusion}
    This research investigated the transferability of adversarial patch attacks from CNN-based object detection to ViT classification models. The experimental evaluation provides several key insights into the vulnerability of state-of-the-art transformer architectures to adversarial attacks.
    The findings demonstrate that both CNN and ViT architectural designs are vulnerable to the same type of attack vector, confirming the cross-architectural transferability of adversarial patches. The results show that patches, originally designed for CNN models, achieved attack success rates of 40.04\% (google/vit-base-patch16-224-in21k) to 99.97\% (facebook/dino-vitb16), with the current facebook/dinov3-vit16 model reaching 65.17\% ASR. This transferability suggests potential for further research into developing unified attack methodologies that target both CNN and ViT models simultaneously, which has particular relevance for mixed pipeline setups where both architectures may be deployed in complementary roles within the same AI pipeline.
    These results highlight vulnerabilities in transformer-based vision systems deployed in security-sensitive applications such as border control and surveillance. Although digital attack effectiveness has been established, further validation is required to assess the approach's performance against detection models such as the latest real-time detection systems (e.g., Roboflow's RF-DETR\footnote{\url{https://roboflow.com/model/rf-detr}}) and to evaluate the attack's effectiveness in physical deployment scenarios under real-world conditions. Future research should focus on physical world validation through comprehensive testing of printed patches under varying lighting conditions, viewing angles, and environmental factors, as well as the development of ViT-specific defense strategies that leverage the unique architectural properties of transformer models. The findings underscore the urgent need for developing robust, architecture-agnostic defense mechanisms to protect critical infrastructure systems that increasingly rely on transformer-based vision models.
    
    While this study focused on demonstrating the cross-architectural transferability of adversarial patches, several promising directions remain for future work, including investigating the role of architectural components such as the Creases Transformation, analysing the impact of patch area and spatial placement on attack effectiveness, and comparing adversarial patches to non-adversarial and random noise baselines to distinguish targeted adversarial behaviour from general perturbation sensitivity. Future research should prioritise physical world validation through comprehensive testing of printed patches under varying lighting conditions, viewing angles, and environmental factors, while simultaneously developing ViT-specific defense strategies that leverage the unique architectural properties of transformer models. These combined efforts would offer deeper insights into the mechanisms driving patch-based vulnerabilities and underscore the urgent need for developing robust, architecture-agnostic defense mechanisms to protect critical infrastructure systems that increasingly rely on transformer-based vision models.
    
    \bibliography{bibliography} 

\begin{thebibliography}{10}

\bibitem{zhang2024MorphingAttack}
Zhang, H., Ramachandra, R., Raja, K., and Busch, C., ``Generalized single-image-based morphing attack detection using deep representations from vision transformer,'' in [{\em Proceedings of the IEEE/CVF Conference on Computer Vision and Pattern Recognition (CVPR) Workshops}{\nolinebreak\hspace{0.1em}]},   1510--1518 (June 2024).

\bibitem{Zhou2025VitAttacks10}
Zhang, C., Zhou, L., Xu, X., Wu, J., and Liu, Z., ``{Adversarial Attacks of Vision Tasks in the Past 10 Years: A Survey},'' {\em ACM Comput. Surv.}  (June 2025).

\bibitem{nguyen2023physical}
Nguyen, K., Fernando, T., Fookes, C., and Sridharan, S., ``Physical adversarial attacks for surveillance: A survey,'' {\em IEEE Transactions on Neural Networks and Learning Systems}  (2023).

\bibitem{brown2017aurko}
Brown, T.~B., Man{\'e}, D., Roy, A., Abadi, M., and Gilmer, J., ``Adversarial patch,'' in [{\em Proceedings of the Advances in Neural Information Processing Systems Workshop}{\nolinebreak\hspace{0.1em}]},  (2017).

\bibitem{5adbb35b1d3a411b8c2ecf9075e9e38b}
Eykholt, K., Evtimov, I., Fernandes, E., Li, B., Rahmati, A., Xiao, C., Prakash, A., Kohno, T., and Song, D., ``Robust physical-world attacks on deep learning visual classification,'' in [{\em Conference on Computer Vision and Pattern Recognition, CVPR 2018}{\nolinebreak\hspace{0.1em}]},   1625--1634 (Dec. 2018).

\bibitem{pmlrv80athalye18b}
Athalye, A., Engstrom, L., Ilyas, A., and Kwok, K., ``Synthesizing robust adversarial examples,'' in [{\em Proceedings of the 35th International Conference on Machine Learning}{\nolinebreak\hspace{0.1em}]},  Dy, J. and Krause, A., eds., {\em Proceedings of Machine Learning Research} {\bf 80},  284--293, PMLR (10--15 Jul 2018).

\bibitem{Thys_2019_CVPR_Workshops}
Thys, S., Van~Ranst, W., and Goedeme, T., ``Fooling automated surveillance cameras: Adversarial patches to attack person detection,'' in [{\em Proceedings of the IEEE/CVF Conference on Computer Vision and Pattern Recognition (CVPR) Workshops}{\nolinebreak\hspace{0.1em}]},  (June 2019).

\bibitem{10.1007/978-3-030-58548-8_1}
Wu, Z., Lim, S.-N., Davis, L.~S., and Goldstein, T., ``Making an invisibility cloak: Real world adversarial attacks on object detectors,'' in [{\em Computer Vision – ECCV 2020: 16th European Conference, Glasgow, UK, August 23–28, 2020, Proceedings, Part IV}{\nolinebreak\hspace{0.1em}]},   1–17, Springer-Verlag, Berlin, Heidelberg (2020).

\bibitem{10.1117/12.3031144}
van Oers, A.~M. and Venema, J.~T., ``{Anti-AI camouflage},'' in [{\em Artificial Intelligence for Security and Defence Applications II}{\nolinebreak\hspace{0.1em}]},  Bouma, H., Prabhu, R., Yitzhaky, Y., and Kuijf, H.~J., eds.,  {\bf 13206},  132060W, International Society for Optics and Photonics, SPIE (2024).

\bibitem{ali2023clip}
Ali, M. and Khan, S., ``Clip-decoder: Zeroshot multilabel classification using multimodal clip aligned representations,'' in [{\em Proceedings of the IEEE/CVF international conference on computer vision}{\nolinebreak\hspace{0.1em}]},   4675--4679 (2023).

\bibitem{radford2021learning}
Radford, A., Kim, J.~W., Hallacy, C., Ramesh, A., Goh, G., Agarwal, S., Sastry, G., Askell, A., Mishkin, P., Clark, J., et~al., ``Learning transferable visual models from natural language supervision,'' in [{\em International conference on machine learning}{\nolinebreak\hspace{0.1em}]},   8748--8763, PmLR (2021).

\bibitem{Ye2024TransformerReIDSurvey}
Ye, M., Chen, S., Li, C., Zheng, W.-S., Crandall, D., and Du, B., ``Transformer for object re-identification: A survey,'' {\em International Journal of Computer Vision} ,  1--31 (2024).

\bibitem{Bayraktar2025ReTrackVLM}
Bayraktar, E., ``Retrackvlm: Transformer-enhanced multi-object tracking with cross-modal embeddings and zero-shot re-identification integration,'' {\em Applied Sciences}~{\bf 15}(4),  1907 (2025).

\bibitem{Khan2022}
Khan, S., Naseer, M., Hayat, M., Zamir, S.~W., Khan, F.~S., and Shah, M., ``{Transformers in Vision: A Survey},'' {\em ACM Comput. Surv.}~{\bf 54} (Sept. 2022).

\bibitem{Pan2022}
Pan, X., Ye, T., Han, D., Song, S., and Huang, G., ``{Contrastive language-image pre-training with knowledge graphs},'' in [{\em Proceedings of the 36th International Conference on Neural Information Processing Systems}{\nolinebreak\hspace{0.1em}]},  {\em NIPS '22}, Curran Associates Inc., Red Hook, NY, USA (2022).

\bibitem{Ebert2023PLGViT}
Ebert, N., Stricker, D., and Wasenmüller, O., ``Plg-vit: Vision transformer with parallel local and global self-attention,'' {\em Sensors}~{\bf 23}(7),  3447 (2023).

\bibitem{Tian2024ViTRobustness}
Tian, R., Wu, Z., Dai, Q., Goldblum, M., Hu, H., and Jiang, Y.-G., ``The role of vit design and training in robustness towards common corruptions,'' {\em IEEE Transactions on Multimedia}  (2024).
\newblock Early Access.

\bibitem{Guesmi2024}
Guesmi, A., Ding, R., Hanif, M.~A., Alouani, I., and Shafique, M., ``{DAP: A Dynamic Adversarial Patch for Evading Person Detectors},'' in [{\em 2024 IEEE/CVF Conference on Computer Vision and Pattern Recognition (CVPR)}{\nolinebreak\hspace{0.1em}]},   24595--24604 (2024).

\bibitem{naseer2021intriguing}
Naseer, M.~M., Ranasinghe, K., Khan, S.~H., Hayat, M., Shahbaz~Khan, F., and Yang, M.-H., ``{Intriguing properties of vision transformers},'' {\em Advances in Neural Information Processing Systems}~{\bf 34},  23296--23308 (2021).

\bibitem{tu2025toward}
Tu, W., Deng, W., and Gedeon, T., ``{Toward a holistic evaluation of robustness in clip models},'' {\em IEEE Transactions on Pattern Analysis and Machine Intelligence}  (2025).

\bibitem{shao2023}
Shao, M., ``{Random Position Adversarial Patch for Vision Transformers},'' (2023).

\bibitem{zhou2023advclip}
Zhou, Z., Hu, S., Li, M., Zhang, H., Zhang, Y., and Jin, H., ``{AdvCLIP: Downstream-agnostic Adversarial Examples in Multimodal Contrastive Learning},'' in [{\em Proceedings of the 31st ACM International Conference on Multimedia}{\nolinebreak\hspace{0.1em}]},  {\em MM '23},  6311–6320, Association for Computing Machinery, New York, NY, USA (2023).

\bibitem{gu2022vision}
Gu, J., Tresp, V., and Qin, Y., ``Are vision transformers robust to patch perturbations?,'' in [{\em European Conference on Computer Vision}{\nolinebreak\hspace{0.1em}]},   404--421, Springer (2022).

\bibitem{Joshi2025}
Joshi, A., Akula, S.~C., Jagatap, G., and Hegde, C., ``A few adversarial tokens can break vision transformers,'' (2025).

\bibitem{zhang2023}
Zhang, J., Huang, Y., Wu, W., and Lyu, M.~R., ``Transferable adversarial attacks on vision transformers with token gradient regularization,'' in [{\em Proceedings of the IEEE/CVF Conference on Computer Vision and Pattern Recognition (CVPR)}{\nolinebreak\hspace{0.1em}]},   16415--16424 (June 2023).

\bibitem{wei2022towards}
Wei, Z., Chen, J., Goldblum, M., Wu, Z., Goldstein, T., and Jiang, Y.-G., ``Towards transferable adversarial attacks on vision transformers,'' in [{\em Proceedings of the AAAI Conference on Artificial Intelligence}{\nolinebreak\hspace{0.1em}]},   {\bf 36}(3),  2668--2676 (2022).

\bibitem{subramanya2022}
Subramanya, A., Saha, A., Koohpayegani, S.~A., Tejankar, A., and Pirsiavash, H., ``{Backdoor Attacks on Vision Transformers},'' (2022).

\bibitem{islam2025}
Islam, C.~M., Chacko, S.~J., Nishino, M., and Liu, X., ``{Mechanistic Understandings of Representation Vulnerabilities and Engineering Robust Vision Transformers},'' (2025).

\bibitem{mahmood2021robustness}
Mahmood, K., Mahmood, R., and Van~Dijk, M., ``On the robustness of vision transformers to adversarial examples,'' in [{\em Proceedings of the IEEE/CVF international conference on computer vision}{\nolinebreak\hspace{0.1em}]},   7838--7847 (2021).

\bibitem{Fawole2025}
Fawole, O. and Rawat, D., ``{Recent Advances in Vision Transformer Robustness Against Adversarial Attacks in Traffic Sign Detection and Recognition: A Survey},'' {\em ACM Comput. Surv.}~{\bf 57} (May 2025).

\bibitem{alam2025adversarial}
Alam, Q.~M., Tarchoun, B., Alouani, I., and Abu-Ghazaleh, N., ``{Adversarial Attention Deficit: Fooling Deformable Vision Transformers with Collaborative Adversarial Patches},'' in [{\em 2025 IEEE/CVF Winter Conference on Applications of Computer Vision (WACV)}{\nolinebreak\hspace{0.1em}]},   7123--7132, IEEE (2025).

\bibitem{ming2024boosting}
Ming, D., Ren, P., Wang, Y., and Feng, X., ``{Boosting the transferability of adversarial attack on vision transformer with adaptive token tuning},'' {\em Advances in Neural Information Processing Systems}~{\bf 37},  20887--20918 (2024).

\bibitem{COCO2014}
Lin, T.-Y., Maire, M., Belongie, S., Hays, J., Perona, P., Ramanan, D., Doll{\'a}r, P., and Zitnick, C.~L., ``Microsoft coco: Common objects in context,'' in [{\em Computer Vision -- ECCV 2014}{\nolinebreak\hspace{0.1em}]},  Fleet, D., Pajdla, T., Schiele, B., and Tuytelaars, T., eds.,  740--755, Springer International Publishing, Cham (2014).

\bibitem{pmlr-v80-athalye18b}
Athalye, A., Engstrom, L., Ilyas, A., and Kwok, K., ``Synthesizing robust adversarial examples,'' in [{\em Proceedings of the 35th International Conference on Machine Learning}{\nolinebreak\hspace{0.1em}]},  Dy, J. and Krause, A., eds., {\em Proceedings of Machine Learning Research} {\bf 80},  284--293, PMLR (10--15 Jul 2018).

\bibitem{Tao2023}
Tao, G., An, S., Cheng, S., Shen, G., and Zhang, X., ``Hard-label black-box universal adversarial patch attack,'' in [{\em 32nd USENIX Security Symposium (USENIX Security 23)}{\nolinebreak\hspace{0.1em}]},   697--714, USENIX Association, Anaheim, CA (Aug. 2023).

\end{thebibliography}
    \bibliographystyle{spiebib} 
    
    \end{document}